\documentclass[lettersize,journal]{IEEEtran}
\usepackage{amsmath,amsfonts}
\usepackage{algorithmic}
\usepackage{algorithm}
\usepackage{array}
\usepackage[caption=false,font=normalsize,labelfont=sf,textfont=sf]{subfig}
\usepackage{textcomp}
\usepackage{stfloats}
\usepackage{url}
\usepackage{verbatim}
\usepackage{graphicx}
\usepackage{cite}
\usepackage{booktabs}
\usepackage{comment}
\usepackage{multicol}
\usepackage{multirow}
\usepackage{makecell}
\usepackage{bbm}
\usepackage{footnote}
\usepackage{enumitem}
\usepackage{amssymb}
\usepackage{xcolor}
\usepackage[colorlinks=true, linkcolor=red, citecolor=blue]{hyperref}
\usepackage{cleveref}

\begin{document}

\title{SIT-FER: Integration of Semantic-, Instance-, Text-level Information for Semi-supervised Facial Expression Recognition}

\author{Sixian Ding*, Xu Jiang*, Zhongjing Du*, Jiaqi Cui, Xinyi Zeng, Yan Wang\dag
\thanks{*: Equal contribution, \dag: Corresponding author}

\thanks{Sixian Ding, Xu Jiang, Zhongjing Du, Jiaqi Cui, Xinyi Zeng, Yan Wang are with the College of Computer Science, Sichuan University, Chengdu 610065, China (E-mail: dingsixian@stu.scu.edu.cn; 2642081986@qq.com; scudzj@qq.com; jiaqicui2001@gmail.com; percyzxy@qq.com; wangyanscu@hotmail.com).}
}
\markboth{Journal of \LaTeX\ Class Files,~Vol.~14, No.~8, August~2021}%
{Shell \MakeLowercase{\textit{et al.}}: A Sample Article Using IEEEtran.cls for IEEE Journals}

\maketitle

\begin{abstract}
Semi-supervised deep facial expression recognition (SS-DFER) has gained increasingly research interest due to the difficulty in accessing sufficient labeled data in practical settings. However, existing SS-DFER methods mainly utilize generated semantic-level pseudo-labels for supervised learning, the unreliability of which compromises their performance and undermines the practical utility. In this paper, we propose a novel SS-DFER framework that simultaneously incorporates semantic, instance, and text-level information to generate high-quality pseudo-labels. Specifically, for the unlabeled data, considering the comprehensive knowledge within the textual descriptions and instance representations, we respectively calculate the similarities between the facial vision features and the corresponding textual and instance features to obtain the probabilities at the text- and instance-level. Combining with the semantic-level probability, these three-level probabilities are elaborately aggregated to gain the final pseudo-labels. Furthermore, to enhance the utilization of one-hot labels for the labeled data, we also incorporate text embeddings excavated from textual descriptions to co-supervise model training, enabling facial visual features to exhibit semantic correlations in the text space. Experiments on three datasets demonstrate that our method significantly outperforms current state-of-the-art SS-DFER methods and even exceeds fully supervised baselines. The code will be available at \url{https://github.com/PatrickStarL/SIT-FER}.

\end{abstract}


\section{Introduction}
\label{sec:intro}

\begin{figure}[t]
  \centering
   \includegraphics[width=1\linewidth]{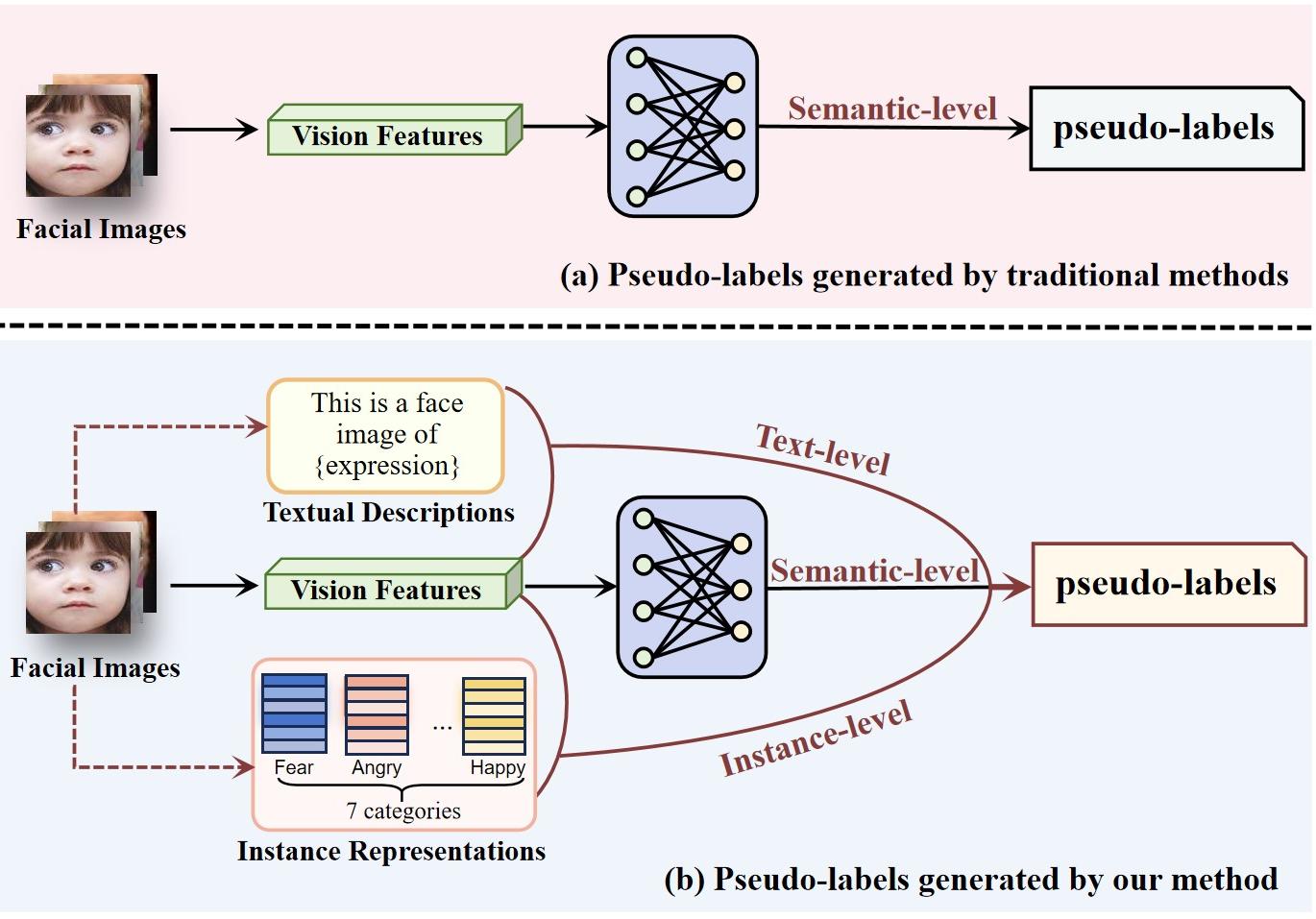}
   \caption{A sketch of the method we propose for generating matching targets: Previous works (a) have only considered similarity at the semantic-level. Since semantic-level information is not always reliable, our method (b) attempts to calculate matching targets by jointly combining instance-, text-, and semantic-level information.}
   \label{fig:intro}
\end{figure}

Facial expressions serve as crucial indicators of human emotions, and their precise recognition is imperative in human-computer interaction. In recent years, the advancement of Deep Facial Expression Recognition (DFER) has been propelled by large-scale annotated datasets \cite{SFEW,RAF-DB,Affectnet}. However, annotating large-scale datasets in real-world scenarios proves to be exceptionally time-consuming and labor-intensive, posing a huge challenge in acquiring high-quality labels.

To make use of the unlabeled data, semi-supervised deep facial expression recognition (SS-DFER) methods have emerged and garnered substantial research interest. For example, Ada-CM \cite{Ada-CM} learns dynamic thresholds for each category to optimize the use of unlabeled data. Meanwhile, LION \cite{Lion} proposes a progressive negative learning module and a relabel mechanism to alleviate the ambiguity in labeled data and avoid information dissipation from unreliable data.

Although current methods have achieved excellent recognition performance, in terms of pseudo-label generation, they still face two major challenges. First, as depicted in Figure \ref{fig:intro}(a), these methods extract semantic-level pseudo-labels from the inherent visual features of facial images \cite{Simmatch}, which are often sharpened to be the final matching targets for strongly enhanced views. Some unreliable labels may inadvertently lead the model in the wrong direction. This problem is exacerbated in SS-DFER because the complexity of face expressions and the high inter-class similarity make traditional semantic-level pseudo-labels even more unreliable. For example, the high similarity between the FEAR and SURPRISE classes makes the pseudo-labels generated for unlabeled images relying only on semantic-level information less confident across classes, and ultimately predicts pseudo-labels that are different from human sensory judgments. We note that the textual information embedded in the labels themselves can provide good a priori information, which when combined with semantic information can produce pseudo-labels with higher confidence.

Second, how to efficiently utilize data has been the focus of semi-supervised learning concerns. In order to solve the limitation of labeled data and little reliable information, we believe that pseudo-labels with very high confidence can be used together with labeled data as a basis for generating pseudo-labels for other unlabeled data. In the field of face expression recognition, data is characterized by high inter-class similarity and intra-class variability. By comparing unlabeled data and all data with high confidence, it can help to find the closest class and supplement information. For the purpose of providing sufficient instance-level information, we construct an instance memory buffer to store the facial vision features and corresponding labels (i.e., one-hot labels and pseudo-labels) for both the labeled and unlabeled data. Notably, for unlabeled data, to ensure the quality of instance information in the memory buffer, we only select samples with reliable predictions to update the buffer while discarding those unreliable ones.

Therefore, we calculate the similarity between the facial visual features and the corresponding textual and instance features respectively and convert them into probabilities, which are then merged with the semantic-level probability to generate the final pseudo-labels, as shown in Figure \ref{fig:intro}(b). In this way, the pseudo-labels can fully integrate the semantic, text, and instance-level information, thus improving the recognition performance.

Moreover, in terms of training, most existing SS-DFER methods mainly focus on model training on labeled data, using visual features extracted from facial images and unique heat label. In order to better utilize the one-hot annotations for the labeled data, we adopt a multimodal supervision strategy that not only utilizes the traditional one-hot labels to constrain the network, but also introduces text embeddings captured from textual descriptions to co-supervise the model training. This empowers the correlation of facial visual features with the text space and enhances the diversity of the supervision signals.

Extensive experiments demonstrate the effectiveness of SIT-FER in various settings. Overall, our main contributions can be summarized as follows:
\begin{itemize}
    \item We propose a novel SS-DFER method SIT-FER, which enables the generation of more reliable pseudo-labels by integrating semantic-, instance- and text-level information. To the best of our knowledge, this is the pioneering work that considers multi-level information in SS-DFER.
    \item We enhance the utilization of one-hot labels by introducing multi-modal supervision signals, which includes facial vision features and text embeddings extracted from label-based textual descriptions. These signals co-supervise the model training, making the facial vision feature posses semantic correlations in text-space.
    \item Extensive experiments on three datasets indicate the effectiveness of the proposed method. Particularly, it sets new records of recognition accuracy with 78.58\% on RAF-DB and 55.14\% on AffectNet, surpassing the second-best state-of-the-art (SOTA) method \cite{Lion} by 2.15\% and 2.43\%.
\end{itemize}

\section{Related Work}
\label{sec:rwork}

\begin{figure*}[htbp]
  \centering
   \includegraphics[clip, trim=0cm 2.4cm 0cm 2.2cm, width=0.95\textwidth]{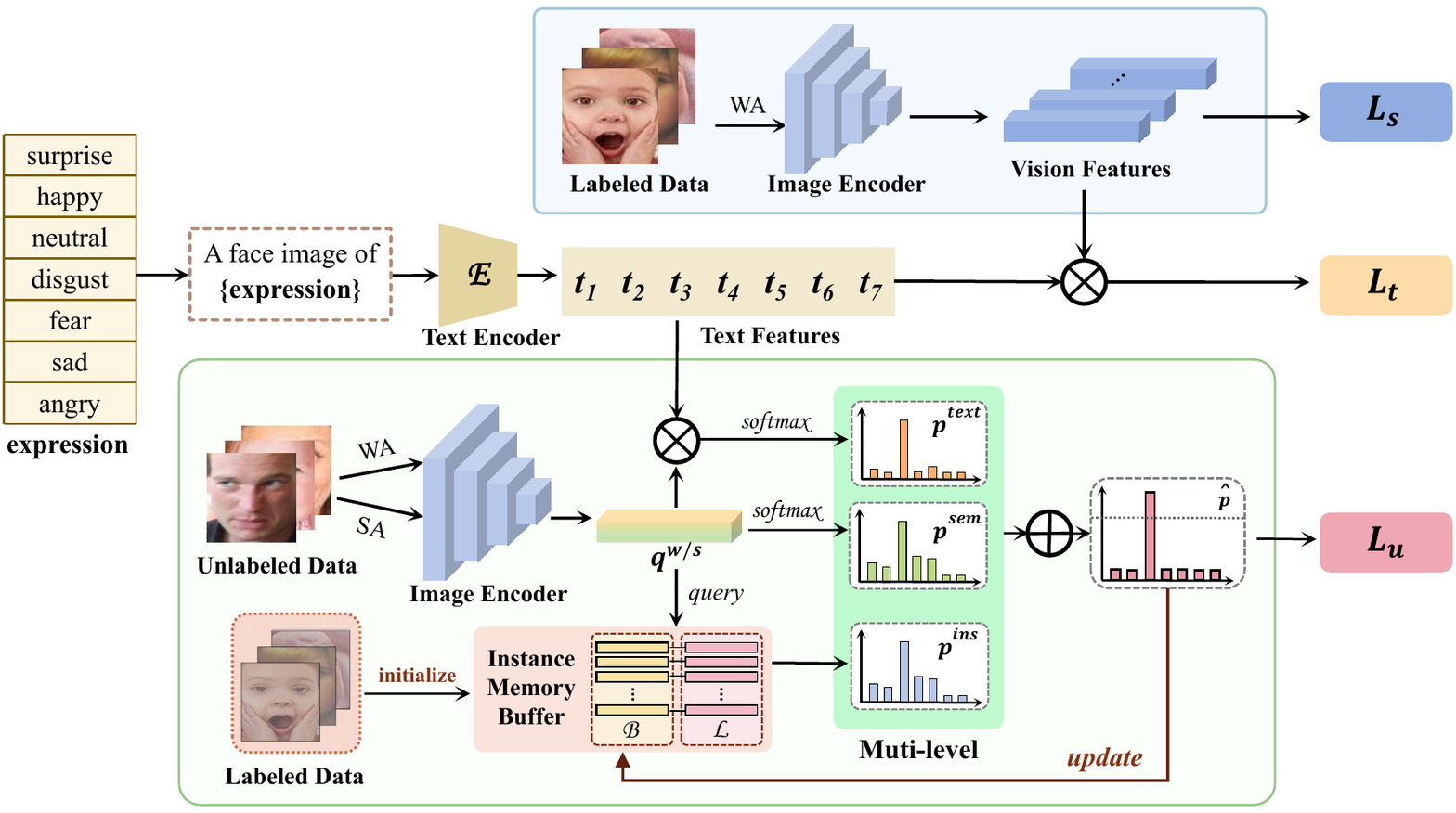}
   \caption{
   Illustration of our  SIT-FER model. $WA$ represents Weak Augmentation and $SA$ represents Strong Augmentation. For labeled data, we utilize the one-hot labels to constrain the network by $L_s$ and calculate the similarity between vision features and text features for co-supervising model training by $L_t$. As for the unlabeled data, we integrate semantic probability, text probability and instance probability as the final pseudo-label $\widehat{p}$ and update the model by the unsupervised consistency loss $L_u$. Additionally, unlabeled data with highly reliable predictions will be added to the Instance Memory Buffer which is initialized by labeled data. Buffer $\mathcal B$ stores the feature embeddings of images and buffer $\mathcal L$ stores their corresponding labels.
   }
   \label{fig:structure}
\end{figure*}

\subsection{Facial Expression Recognition}
Facial Expression Recognition (FER) presents a challenge for computers to accurately discern and interpret human emotions. Previous studies such as CK+ \cite{ck+} and Oulu-Casia \cite{Oulu-Casia} primarily relied on laboratory-made datasets. However, real-world images often suffer from uneven quality and inaccurate annotations, posing challenges in acquiring high-fidelity datasets comparable to controlled laboratory environments. In recent years, an increasing number of researchers \cite{wang2020graph,she2021dive,Transfer,Choi2023DCNN,FER-former,yang2023facial,Lv2024VTAnet} have turned their attention to in-the-wild datasets to address the problems of fuzzy labels and false semantic information. Zeng et al. \cite{zeng2018facial} introduced multiple training phases to solve the problem of annotation inconsistencies. Chen et al. \cite{chen2020label} constructed an auxiliary label space diagram to explore label distribution. Wang et al. \cite{wang2020region} focused on mining the potential truth value and confidence weight of each sample to reduce the influence of ambiguous data. Li et al. \cite{li2022deep} proposed a deep margin-sensitive representation learning framework for cross-domain facial expression recognition. Liu et al. \cite{Liu2024MTAC} addressed data uncertainty through confidence estimation, auxiliary tasks, and feature-based re-labeling. She et al. \cite{she2021dive}, which reaches the current SOTA performance, explored the potential distributions in the label space and estimated pair uncertainty.

However, achievements in in-the-wild FER obtained so far are based on complete supervision and rely on a large number of high-quality manual labeling, which is time-cost and labor-intensive. To reduce the dependency on exhaustive manual annotation, Margin-mix method \cite{Margin-mix} applied SS-DFER for the first time and then an extension of Mixmatch \cite{Mixmatch} is proposed. Li et al. \cite{Ada-CM} proposed an adaptive threshold approach to generate reliable pseudo-labels for unlabeled samples with high confidence. Du et al. \cite{Lion} introduced a progressive negative learning module and a relabel module to resolve the issues of ambiguous labels in labeled data and avoid wasting information from unreliable data. Zhang et al. \cite{Zhang2022WeakSupervised} proposed a pose-invariant model that leverages noisy web data and introduces pose and noise modeling networks to enhance performance. Currently, SS-DFER remains an emerging field with great research potential. We note that generating pseudo labels only by semantic-level probability is not always reliable, and can lead to model misinterpretation. In our work, we integrate information from the semantic-, instance- and text-level to alleviate the unreliability of traditional semantic-level pseudo-labels in SS-DFER.

\subsection{Semi-Supervised Learning}
Semi-Supervised Learning (SSL) allows for the efficient utilization of both limited labeled data and abundant unlabeled data, thereby alleviating the burden of manual labeling while guaranteeing the model performance at the same time. At present, SSL methods can be divided into five main categories: graph-based methods \cite{marino2016more,wang2020graph,Lu2024Graph}, generation model-based methods \cite{donahue2016adversarial,denton2016semi}, methods using consistent regularization \cite{sajjadi2016regularization,xie2020unsupervised}, pseudo notation methods \cite{rizve2021defense,pham2021meta} and hybrid methods \cite{Fixmatch,Flexmatch,Dash,Hu2025multi}.

Among them, the hybrid methods usually perform better than other methods on various datasets, including CIFAR-10, CIFAR-100 \cite{krizhevsky2009learning} and ImageNet \cite{ImageNet}. For example, Xie et al. \cite{xie2020unsupervised} and Sohn et al. \cite{Fixmatch} set thresholds when exporting pseudo-labels for weakly augmented views of unlabeled data, and then use the generated pseudo labels to supervise predictions for strongly augmented views. Zhang et al. \cite{Flexmatch} and Xu et al. \cite{Dash} further applied the dynamic threshold method to semi-supervised tasks. However, due to the unreliability of traditional pseudo-labels, direct applications of these SSL methods to SS-DFER tasks do not yield satisfactory results. In our work, we propose a Multi-level Information Pseudo-label Generation method to generate reliable pseudo labels for better guiding the model learning.

\section{Method}
\label{sec:method}
Our proposed SIT-FER aims to tackle the Facial Expression Recognition (FER) task in a semi-supervised setting. Specifically, We introduce multi-modal supervisory signals to jointly supervise model learning, thereby alleviating label ambiguity issues. Additionally, we propose a method for generating more reliable pseudo-labels by integrating semantic-, instance-, and text-level information to better utilize unlabeled data. In this section, we first outline our problem formulation and provide an overview of our method in Sec. \ref{overview}. Then, the multimodal supervisory signals are introduced in Sec. \ref{Multi-Supervise}. Furthermore, we detailedly describe the generation of more reliable pseudo-labels in Sec. \ref{Multi-generate}. After that, we elaborate on the details of the Instance Memory Buffer in Sec. \ref{buffer}. Finally, we formulate the whole training objective in Sec. \ref{allfunctions}.

\subsection{Overview}
\label{overview}
We define the semi-supervised image classification problem as follows: Given a batch of $B$ labeled samples $X={x_b:b\in(1,\ldots,B)}$ with their labels $Y={y_b:b\in(1,\ldots,B)}$ and a batch of $\mu B$ unlabeled samples $U={u_b:b\in(1,\ldots,\mu B)}$, our goal is to train a robust model to accurately distinguish different expressions. Figure \ref{fig:structure} provides an overview of our model. For labeled data, we first apply Weak Augmentation (WA) and then use an image encoder to obtain vision feature vectors $f$, followed by a fully connected class prediction head to derive its prediction probability $p$. Given a dataset containing $K$ ($K=7$) categories of expressions, we generate a text label vector for each class, represented as $T={T_1, T_2, \ldots, T_K}$. We then use a text encoder to extract text features $t={t_1, t_2, \ldots, t_k}$ for calculating the similarity between image features and text features, along with the text-level prediction probability. The calculated text and image-level prediction probabilities are used to calculate cross-entropy loss, thereby enhancing the supervised learning process. Moreover, an Instance Memory Buffer is introduced to store feature embeddings and corresponding labels of all the labeled instances, offering instance-level information support during the processing of unlabeled data. For unlabeled images, both WA and Strong Augmentation (SA) are applied, and the weight-sharing feature extractor is employed to obtain the corresponding feature vectors $q^w$, $q^s$, and their prediction probabilities $p^{sem}$ and $p^s$, respectively.

Unlike previous methods that rely solely on semantic probability as pseudo-labels, our pseudo-label generation also integrates text space probability $p^{text}$ and instance space probability $p^{ins}$ to formulate the final pseudo-label $\widehat{p}$. If $\max \widehat{p}$  is greater than a certain threshold, it will be used to supervise the SA version via the cross-entropy loss and update the Instance Memory Bank. We will provide more details in the following sections.
\begin{figure*}[t]
  \centering
   \includegraphics[width=0.9\linewidth]{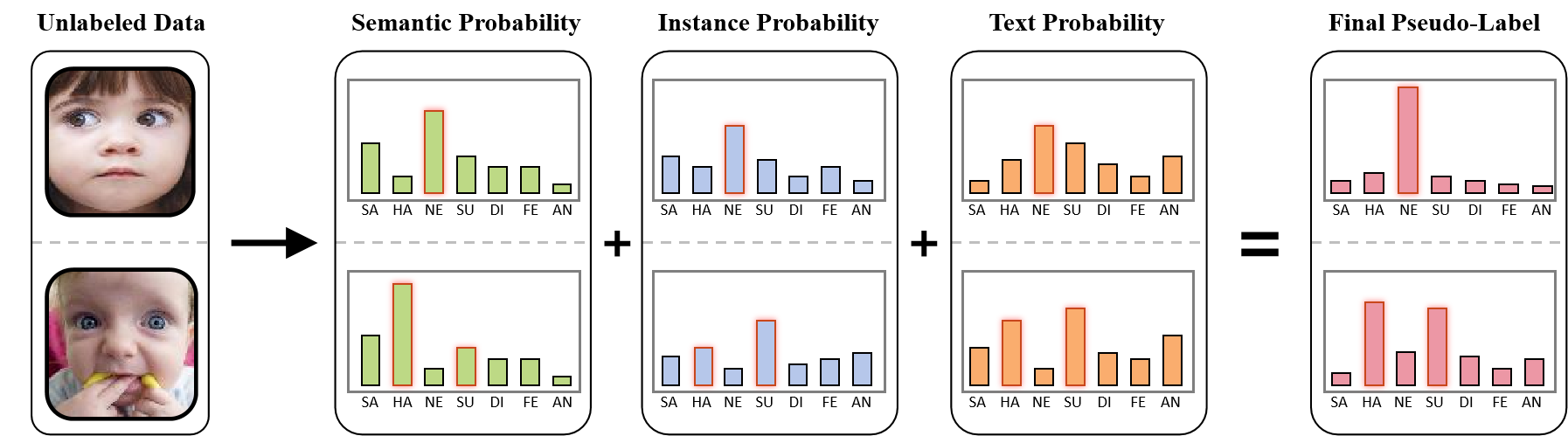}
   \caption{Diagram of our Multi-level Information Pseudo-label Generation method processing clear and ambiguous samples. If the predictions from semantic-, instance- and text-level are similar, the result pseudo-label will be much sharper and produce high confidence for some classes. When these three predictions are different, the result pseudo-label will be much smoother.
    }
\label{fig:result}
\end{figure*}

\subsection{Multimodal Supervisory Signals}
\label{Multi-Supervise}
Given labeled data $X$ with their labels $Y$, we first apply WA function $T_w (\cdot)$ and use an encoder $F(\cdot)$ to extract vision features from these WA views, i.e., $f=F(T_w(x))$. Then, a fully connected class prediction head $\phi(\cdot)$ is used to map \(f\) into semantic probabilities, which is written as $p=\phi(f)$. The one-hot label $y$ can be used to supervise the probability $p$ through cross-entropy loss as follows:
\begin{equation}
\label{eq:ls}
    L_s = \frac{1}{B} \sum CE(y, p)
\end{equation}
where $CE(\cdot)$ denotes the standard cross-entropy.

In addition, given a dataset contains $K\ (K=7)$ expressions, text labels can be expressed as 'a face image of \{expression\}', where \{expression\} signifies the corresponding class in [surprise; fear; disgust; happy; sad; angry; neutral]. We utilize a distinct text encoder to extract text features $t$. Then, the similarity between image features $f_j$ (i.e., $j$ represents the $j^{th}$ image) and text features $t$ is calculated using a similarity function $sim(\cdot)$:
\begin{equation}
\label{eq:sb-t}
    S_{j}^{text} = sim(f_j, t) =\frac{f_j \cdot t^T}{\|f_j\| \cdot\|t\|}
\end{equation}

A softmax function is then adopted to process the calculated similarities as follows:
\begin{equation}
\label{eq:ps}
    P_{{S}_{j}^{text}} = \frac{e^{{{S}_{j}^{text}}/{\tau}}}{\sum_{b=1}^B \left({e^{{{S}_{b}^{text}}/{\tau}}}\right)}
\end{equation}
where $ \tau $ is the temperature parameter that controls the sharpness of the distribution. Finally, the text-level supervision can be achieved by cross entropy loss, which can be formulated as:
\begin{equation}
\label{eq:lt}
    L_t = -\sum_{b=1}^B {y_b\log({P}_{S_{b}^{text}})}
\end{equation}

\begin{algorithm}[t]
    \caption{The training process of SIT-FER}
    \label{alg}
    \begin{algorithmic}[1]
        \renewcommand{\algorithmicrequire}{\textbf{Input:}}
        \REQUIRE Labeled set $X=\{x_b:b\in(1,\ldots,B)\}$, unlabeled set $U=\{u_b:b\in(1,\ldots,\mu B)\}$.
        \renewcommand{\algorithmicrequire}{\textbf{Parameters:}}
        \REQUIRE Maximum number of epochs $E$, thresholds $\tau$ and $\gamma$.
        \renewcommand{\algorithmicensure}{\textbf{Output:}}
        \ENSURE Updated SIT-FER model.
        
        \STATE  /*Training*/
         \FOR{$e=1:E$}
             \STATE  /*For labeled set*/
             \FOR{$b=1:B$}
                \STATE Compute $L_s$ by Eq.\ref{eq:ls}.
                \STATE Extract text features $t$ and calculate text-image similarity by Eq.\ref{eq:sb-t}.
                \STATE Derive text-level prediction probability $p^{text}$ and compute $L_{t}$ by Eq.\ref{eq:lt}.
                \STATE /* Update Instance Memory Buffer */
                \STATE Update Buffer with features of $x_b$ and the label.
            \ENDFOR

            \STATE  /*For unlabeled set*/
            \FOR{$b=1:\mu B$}
                \STATE Weakly and strongly augment $u_b$ and obtain its  feature vectors $q^w$ and $q^s$ as well as its prediction probabilities $p^{sem}$ and $p^s$. 
                \STATE Compute text-image similarity and text-level probability $p^{text}$ by Eq.\ref{eq:ptext}.
                \STATE Obtain instance embeddings for $q^w$ and calculate instance-level probability $p^{ins}$ by Eq.\ref{eq:si}, \ref{eq:pi}.
                \STATE Generate final pseudo-label $\widehat{p}$ integrating semantic, text, and instance probabilities by Eq.\ref{eq:phat}.
                
                \IF {$\max{\widehat{p}}\geq \alpha$}
                    \STATE Calculate $L_u$ by Eq.\ref{eq:lossu}.
                    \IF {$\max{\widehat{p}}\geq \gamma$}
                        \STATE Update the instance memory buffer with features of $u_b$ and pseudo-label.
                    \ENDIF
                \ENDIF
            \ENDFOR
        \ENDFOR       
    \end{algorithmic}
\end{algorithm}

\subsection{Multi-level Information Pseudo-label Generation}
\label{Multi-generate}

Prior works typically use the output vectors of a fully connected layer as pseudo-labels. However, it is not always dependable, particularly for SS-DFER tasks due to the high inter-class similarity and scarcely labeled data. To address this problem, we propose a novel method for generating more reliable pseudo-labels by integrating text-, instance- and semantic-level information in this work. This approach improves the utilization efficiency of unlabeled data and enhances the 
performance of the model by creating more credible criteria for strongly augmented views.

For each unlabeled sample, we use the same processing step as the labeled samples to obtain the features $q^w$ and $q^s$ corresponding to the weak and strong augmented views of the unlabeled data. Similar to the labeled samples, $\phi(\cdot)$ is used to map $q^w$ and $q^s$ into semantic similarities, \textit{i.e.}, $p^{sem}=\phi(q^w)$ and $p^s = \phi(q^s)$.

For text-level information, we compute the probability between image features and the text space, which can be formulated as:
\begin{equation}
\label{eq:ptext}
    p^{text} = softmax\left({sim\left(q^w, t\right)}/{\tau}\right)
\end{equation}

In addition to semantic- and text-level probabilities, we also consider instance-level probabilities. Suppose we have a non-linear projection head $g(\cdot)$ that maps the representation $q_i$ (i.e. $q_i$ denotes the feature of the $i^{th}$ unlabeled image) to a low-dimensional embedding $z_i=g(q_i)$. The feature embedding of the WA view can be represented as $z_i^w=g(q_i^w)$. As mentioned in Sec. \ref{overview}, an Instance Memory Buffer storing the feature embeddings of $K$ instance samples is constructed, denoted as $\{z_m:m\in(1,\ldots,M)\}$, where $M$ equals the number of labeled samples. We calculate the similarity between the image and the $j^{th}$ instance in the Instance Memory Buffer (details described in Sec. \ref{buffer}) as follows:
\begin{equation}
\label{eq:si}
    S_j = \frac{e^{sim(z_i^w, z_j)/{\tau}}}{\sum_{m=1}^M e^{sim(z_i^w, z_m)/{\tau}}}
\end{equation}
where global instance similarity $\{S_j:j\in(1,\ldots,M)\}$ encompasses the similarities of all instances across all classes. Considering the intra-class variability in facial expression recognition, we encourage the model to focus more on the local instance similarity within each class. To involve instance similarity in the final pseudo-label generation, we extract the $K$-dimensional local similarity $p^{ins}$ from the $M$-dimensional global instance similarity $S_j$:
\begin{equation}
\label{eq:pi}
    p_i^{ins}=\max{\{\mathbbm{1}(class(S_j)=i)S_j,j\in(1,\ldots,M)\}}
\end{equation}
where $class(\cdot)$ is the function that returns the ground truth expression category. Specifically, $class(S_j)$ indicates the label for the $i^{th}$ instance in the Instance Memory Buffer. We can now generate the final pseudo-label $\widehat{p}$ \,based on the summation of semantic similarity $p^{sem}$, text similarity $p^{text}$, and instance similarity $p^{ins}$, which can be expressed as:
\begin{equation}
\label{eq:phat}
    \widehat{p} = \frac{p^{sem} + p^{text} + p^{ins}}{3}
\end{equation}
As such, the final pseudo-label $\widehat{p}$ incorporates semantic-, text-, and instance-level information. Our insight is that, if the probabilities at the three levels are divergent, the resulting pseudo-labels will be smoother, as shown in Figure \ref{fig:result}. Conversely, the pseudo-labels will contain higher probability values. Ultimately, we harness the pseudo-labels to supervise the prediction of the SA views (represented by $p^s$) similar to FixMatch \cite{Fixmatch}:
\begin{equation}
    \label{eq:lossu}
    L_u = \frac{1}{\mu B} \sum_{b=1}^{\mu B} \mathbbm{1}(\max DA(\widehat{p}) > \alpha)CE(DA(\widehat{p}), p^s)
\end{equation}
where $\alpha$ is the confidence threshold. Following previous methods \cite{Fixmatch}, the model selects out samples where the maximum pseudo-label class probability exceeds the confidence threshold. This approach ensures that only samples with a high degree of confidence in their pseudo-labels are considered, thus enhancing the overall reliability of the model. $ DA(\cdot)$ represents the distribution alignment strategy in \cite{Remixmatch}, which balances the distribution of pseudo-labels. In the calculation of the unsupervised loss,\, $\widehat{p}$ \, is a soft label and is not sharpened into a one-hot label. Following \cite{Comatch}, we maintain a moving average of $\widehat{p}_{avg}$ and adjust the current $\widehat{p}$ with $Normalize(\widehat{p}/\widehat{p}_{avg})$.

\subsection{Instance Memory Buffer}
\label{buffer}
In the process of handling unlabeled data, an Instance Memory Buffer is constructed to store feature embeddings and corresponding labels of all labeled instances, thereby providing instance information for the prediction of unlabeled data. Specifically, we have established two global buffers, \textit{i.e.,} $B\in {\mathbb{R}}^{M\times S}$ that stores the feature embeddings of images and $L\in {\mathbb{R}}^{M\times 1}$ that stores their corresponding labels. $M$ is the number of labeled data, and $S$ represents the dimension of the feature embedding.
For the update of the Instance Memory Bank, we adopt a temporal integration strategy \cite{Pseudo-label,wu2018unsupervised} to smoothly update the features in memory, which can be written as follows:
\begin{equation}
\label{eq:zt}
    z_t \leftarrow mz_t + (1 - m)z_{t-1}
\end{equation}
where $z_t$ is the feature in the Instance Memory Buffer for the $t^{th}$ epoch. It is worth noting that during the training process, unlabeled data with highly reliable predictions will be added to the Instance Memory Buffer. Specifically, if the final pseudo-label of an unlabeled image meets the condition $\max{DA(\widehat{p_i})}>\gamma$, it can be considered highly reliable. Here, $\gamma$ is a relatively high confidence threshold and its value will be greater than $\alpha$. Then its corresponding feature embedding will be used to expand the Instance Memory Buffer by concatenating $B$ and $z_i$. The predicted category $class(\widehat{p_i})$ corresponding to $z_i$ will also be added to $L$.

\begin{table*}[t]
    \caption{Performance comparison (\%) with the state-of-the-art methods on RAF-DB, SFEW and AffectNet. The best results are in bold font and the second-best results are underlined.}
    \centering
    \vspace{-0.2cm}
    \renewcommand\arraystretch{1.2}
    \resizebox{1.0\linewidth}{!}{
        \begin{tabular}{lcccccccc}
    	\toprule
    	\multirow{2}{*}{~~~~~~~\textbf{Method}}&
		  \multicolumn{4}{c}{\textbf{RAF-DB}}&
            \multicolumn{2}{c}{\textbf{SFEW}}& 
    	\multicolumn{2}{c}{\textbf{AffectNet}}\cr
    	\cmidrule(lr){2-5}  \cmidrule(lr){6-7}  \cmidrule(lr){8-9}
    		 &100 labels &400 labels &2000 labels & 4000 labels &100 labels &400 labels &2000 labels &10000 labels\\
    		
            \midrule
            Baseline & 52.43±\footnotesize{2.24} & 67.75±\footnotesize{0.95} & 78.91±\footnotesize{0.43} & 81.90±0.48 & 33.76±1.84 & 43.85±2.83 & 47.52±0.75 & 53.18±0.68 \\
            Pseudo-Labeling \cite{Pseudo-label} & 54.96±\footnotesize{4.24} & 69.99±\footnotesize{1.81} & 79.18±\footnotesize{0.27} & 82.88±\footnotesize{0.49} & 34.27±\footnotesize{1.67} & 45.27±\footnotesize{1.32} & 48.78±\footnotesize{0.67} & 53.82±\footnotesize{1.29} \\
            MixMatch \cite {Mixmatch} & 54.57±\footnotesize{4.16} & 73.14±\footnotesize{1.40} & 79.63±\footnotesize{0.91} & 83.57±\footnotesize{0.49} & 34.13±\footnotesize{2.58} & 44.91±\footnotesize{1.87} & 49.63±\footnotesize{0.49} & 53.49±\footnotesize{0.47} \\
            UDA \cite{UDA} & 58.15±\footnotesize{1.54} & 72.39±\footnotesize{1.64} & 81.16±\footnotesize{0.54} & 83.56±\footnotesize{0.82} & 39.22±\footnotesize{2.30} & 48.90±\footnotesize{1.56} & 50.42±\footnotesize{0.45} & 56.49±\footnotesize{0.27} \\
            ReMixMatch \cite{Remixmatch} & 58.83±\footnotesize{2.34} & 73.34±\footnotesize{1.82} & 79.66±\footnotesize{0.66} & 83.51±\footnotesize{0.18} & 35.69±\footnotesize{2.73} & 48.39±\footnotesize{0.71} & 50.38±\footnotesize{0.63} & 55.81±\footnotesize{0.34} \\
            MarginMix \cite{Margin-mix} & 58.91±\footnotesize{1.78} & 73.31±\footnotesize{1.64} & 80.22±\footnotesize{0.76} & 83.47±\footnotesize{0.28} & 38.69±\footnotesize{1.93} & 49.21±\footnotesize{0.92} & 50.58±\footnotesize{0.42} & 56.41±\footnotesize{0.28} \\
            FixMatch \cite{Fixmatch} & 60.67±\footnotesize{2.25} & 73.36±\footnotesize{1.59} & 81.27±\footnotesize{0.27} & 83.31±\footnotesize{0.33} & 38.90±\footnotesize{1.90} & 50.73±\footnotesize{0.45} & 50.79±\footnotesize{0.37} & 56.50±\footnotesize{0.43} \\
            Ada-CM \cite{Ada-CM} & 62.36±\footnotesize{1.10} & 74.44±\footnotesize{1.53} & 82.05±\footnotesize{0.22} & 84.42±\footnotesize{0.49} & 41.88±\footnotesize{2.12} & 52.43±\footnotesize{0.67} & 51.22±\footnotesize{0.29} & 57.42±\footnotesize{0.43} \\
            LION \cite{Lion} & \underline{67.83±\footnotesize{0.64}} & \underline{76.43±\footnotesize{1.12}} & \underline{82.39±\footnotesize{0.13}} & \underline{84.81±\footnotesize{0.16}} & \underline{45.61±\footnotesize{0.32}} & \underline{54.18±\footnotesize{0.52}} & \underline{52.71±\footnotesize{0.21}} & \underline{59.11±\footnotesize{0.38}} \\
    		
            \midrule
            SIT-FER & \textbf{69.75±\footnotesize{1.84}} & \textbf{78.58±\footnotesize{1.43}} & \textbf{83.12±\footnotesize{0.11}} & \textbf{84.93±\footnotesize{0.16}} & \textbf{46.48±\footnotesize{1.94}} & \textbf{55.76±\footnotesize{0.81}} & \textbf{55.14±\footnotesize{0.53}} & \textbf{60.87±\footnotesize{0.29}} \\
            
            \cmidrule(lr){2-5}  \cmidrule(lr){6-7}  \cmidrule(lr){8-9}
            Fully Supervised & \multicolumn{4}{c}{84.13} & \multicolumn{2}{c}{51.05} & \multicolumn{2}{c}{52.97} \\
            \bottomrule
        \end{tabular}
    }
    \label{cmp}
\end{table*}

\begin{table}[t]
\centering
    \caption{Evaluation (\%) of MIPG ($p^{text}$ and $p^{ins}$) and $L_{t}$ on RAF-DB and SFEW.}
    \vspace{-0.2cm}
    \begin{tabular}{p{0.6cm}<{\centering} p{0.6cm}<{\centering} p{0.6cm}<{\centering} p{0.6cm}<{\centering} p{1.8cm}<{\centering} p{1.8cm}<{\centering}}
    \toprule
    \multicolumn{3}{c}{MIPG} & \multirow{2}{*}{\textit{$L_{t}$}} & RAF-DB & SFEW \cr
    \cmidrule{1-3}  \cmidrule{5-6}
    $p^{sem}$ & $p^{text}$ & $p^{ins}$ & & 100 labels & 400 labels  \\
    \midrule
    \checkmark & & & &  61.38±\footnotesize{1.38} & 51.13±\footnotesize{1.46}  \\
    \checkmark & \checkmark & & & 63.09±\footnotesize{1.03} & 52.41±\footnotesize{1.83}  \\
    \checkmark & \checkmark & \checkmark & & 65.24±\footnotesize{1.55} & 54.11±\footnotesize{1.98} \\
    \checkmark & \checkmark & \checkmark & \checkmark & 69.75±\footnotesize{1.84} & 55.76±\footnotesize{0.81} \\
    \bottomrule
    \end{tabular}
    \label{evl}
    \vspace{-0.2cm}
\end{table}

\subsection{Overall Objective Function}
\label{allfunctions}
As mentioned above, there are three losses to optimize the parameters of our SIT-FER model: 1) $L_s$ loss on labeled data; 2) $L_t$ loss between image and text pairs; 3) $L_u$ on unlabeled data. The total loss is formulated as follows: 
\begin{equation}
    L_{total}=L_s+\lambda_1 L_t+\lambda_2 L_u
\end{equation}
where $\lambda_1$ and $\lambda_2$ are hyper-parameters to balance these terms.

\section{Experiments}
\label{sec:exp}

\subsection{Datasets and Metrics}
\label{datasets}
\quad \textbf{Datasets.} We validate the efficacy of our SIT-FER model across three datasets, including: RAF-DB \cite{RAF-DB}, AffectNet \cite{Affectnet}, and SFEW \cite{SFEW}. \textbf{RAF-DB} encompasses 30,000 facial images, each meticulously annotated by a team of 40 specialists. For our study, we pinpoint seven emotions, \textit{i.e.,} fear, surprise, sadness, happiness, disgust, anger, and neutral, as our classification targets. The dataset is bifurcated into a training set of 12,271 images and a test set comprising 3,068 images. \textbf{AffectNet }is a more expansive dataset with a repertoire of 420,000 facial images, each manually labeled with one of eight expressive tags. Similar to the category selection of RAF-DB, our study harnesses the identical seven emotions, utilizing 240,000 images for training purposes and earmarking a batch of 3,500 images for testing. \textbf{SFEW} is composed of static cinematic frames extracted from movies, including 958 training images and 436 test images. To authentically simulate a semi-supervised scenario, we intentionally omit labels on a randomized selection of images at varying proportions.

\textbf{Performance Metrics.} We conduct experiments employing multiple random seeds and compute the average accuracy as well as the standard deviation on the test set to assess the performance of the method.

\subsection{Implementation Details}
\label{implement}
By default, ResNet-18 is used as backbone network for the image encoder, which is pre-trained on MS-Celeb-1M face recognition dataset \cite{Ms-celeb-1m}. CLIP-ViT-B-32 is adopted as the backbone network for the text encoder, which is pre-trained on a diverse range of internet-collected data for robust multimodal understanding \cite{CLIP}. Facial images are aligned and resized to $224\times224$ by MTCNN \cite{MTCNN}. RandomCrop and RandomHorizontalFlip are employed as weak augmentation. RandAugment \cite{Randaugment} is used as strong augmentation following \cite{Ada-CM}. The whole network is trained for 60 epochs with the Adam optimizer. The initial learning rate is set to \( 5 \times 10^{-4} \). The batch size is 16. The above setting keeps consistent with that of all the compared methods for fairness. As for the hyperparameters, we set \(\tau=0.80\), \(\gamma=0.86\) and the trade-off parameters \(\lambda_1\), \(\lambda_2\) are set to $0.6$, $0.3$, respectively.

\subsection{Comparison With the State-of-the-Art}
\label{comparison}
To verify the performance of SIT-FER, we compare it with a range of state-of-the-art methods, including Pseudo-Labeling \cite{Pseudo-label}, MixMatch \cite{Mixmatch}, UDA \cite{UDA}, Margin-Mix \cite{Margin-mix}, ReMixMatch \cite{Remixmatch}, FixMatch \cite{Fixmatch},  Ada-CM \cite{Ada-CM} and LION \cite{Lion}, on all the three datasets with different ratios of labeled data. All these comparing methods represent substantial and impactful contributions to semi-supervised image classification. We have tailored them to suit our SS-DFER task. Particularly, we regard our model trained using only limited labeled data as the baseline. Besides, we apply DLP-CNN \cite{DLPCNN} on RAF-DB and SFEW, RAN \cite{RAN} on AffectNet as fully-supervised baseline.

Table \ref{cmp} presents the comparative results. As can be seen, all semi-supervised strategies outperform the baseline due to their utilization of unlabeled data. Compared to the second-best LION method, our approach outperforms it with a large margin across all datasets and all ratio of labeled data. Even with a mere 100 labeled images in RAF-DB and SFEW, our method still leads with accuracies of 69.75\% and 46.48\%, respectively, demonstrating a potent capability to leverage unlabeled data.
Moreover, our method can even beat fully-supervised methods with 4,000, 400, and 10,000 labeled instances for RAF-DB, SFEW, and AffectNet, respectively, achieving margins of 0.80\%, 4.71\%, and 7.90\%. 

\subsection{Ablation Study}
\label{ablation}

In this section, we conduct an ablation study to verify the contributions of the key components and explore the optimal values of hyperparameters in SIT-FER.

\textbf{Effectiveness of Components in SIT-FER.} In our SIT-FER model, there are two critical components: 1) Multi-modal Signal Supervised Learning (MSSL, for short) and 2) Multi-level Information Pseudo-label Generation (MIPG, for short). We isolate them from the entire module to evaluate their contributions. Specifically, we attribute the contribution of multi-modal signal supervised learning to the \(L_{t}\) loss. We establish a baseline by removing the \(L_{t}\) loss, the \(p^{text}\) and \({p^{ins}}\) within the MIPG module. Experiments with 100 and 400 labels were conducted on RAF-DB and SFEW respectively, and the results are presented in Table \ref{evl}. As can be observed, the MIPG module significantly enhances performance by 3.86\% on RAFDB and by 2.98\% on SFEW, demonstrating its ability to generate more reliable pseudo-labels, thereby enhancing the efficiency of utilizing unlabeled data. Meanwhile, we can find that both text probability \(p^{text}\) and instance probability \(p^{ins}\) can better complement semantic probability  contributing performance improvements of 1.71\% for \(p^{text}\) and 2.15\% for \(p^{ins}\) on RAF-DB, and 1.28\% for \(p^{text}\) and 1.70\% for \(p^{ins}\) on SFEW. Moreover, when \(L_{t}\) is incorporated, the performance further increases by 4.51\% on RAF-DB and 1.65\% on SFEW. Furthermore, it can be seen that combining the loss simultaneously with the MIPG module results in even greater performance gains, as both modules actually complement each other. In summary, all these results fully demonstrate the effectiveness of the proposed components.

\begin{figure}[t]
  \centering
   \includegraphics[width=0.7\linewidth]{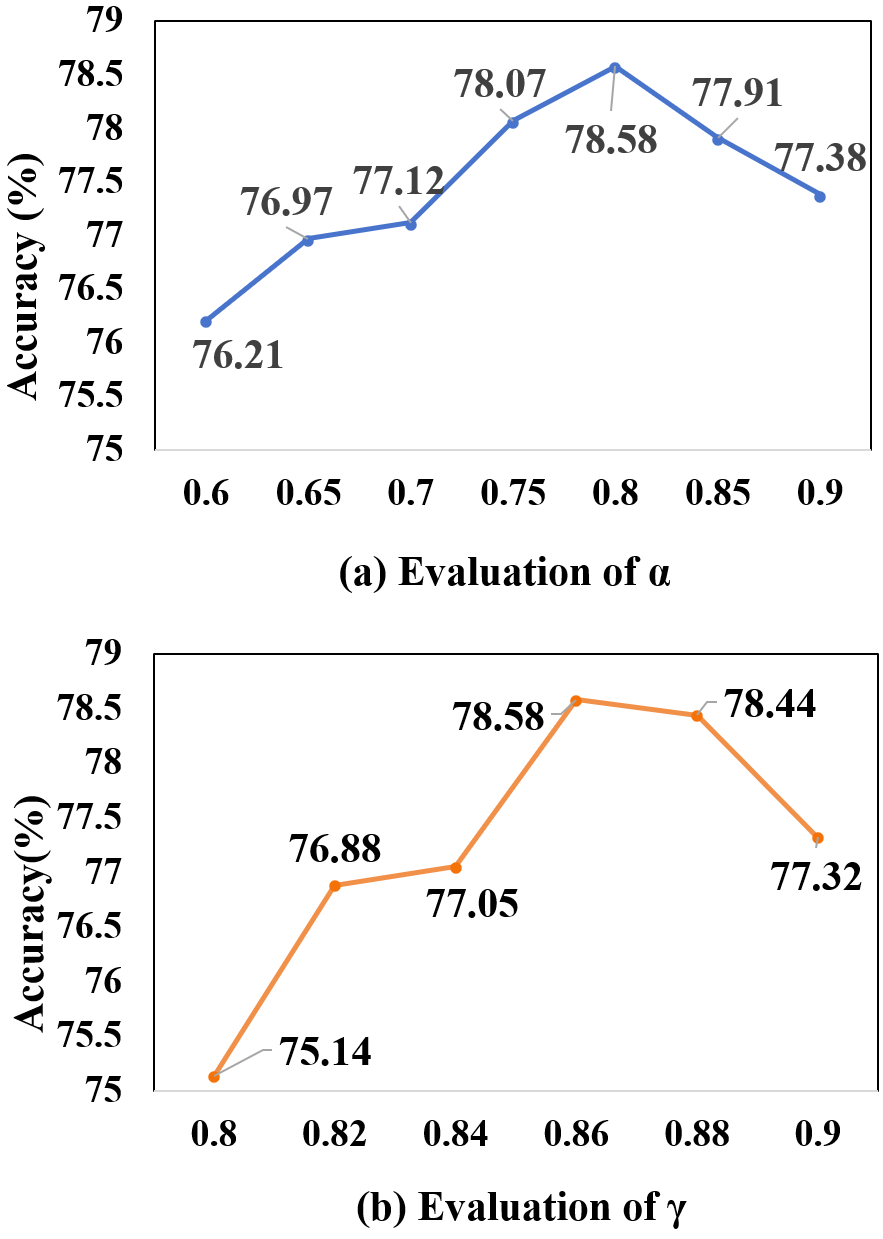}
   \vspace{-0.3cm}
   \caption{Evaluation of parameters $\alpha$ and $\gamma$ on RAF-DB.}
   \label{fig:parameter}
\end{figure}

\begin{table}[t]
    \caption{Evaluation(\%) of different text contents on RAF-DB and SFEW datasets.}
    \vspace{-0.2cm}
\centering
        \begin{tabular}{p{1.4cm}<{\centering} p{4.8cm}<{\centering} p{1.2cm}<{\centering}}
        \toprule
        Datasets                           & Texts contents                         & Acc.(\%)    \\
        \midrule
        \multirow{3}{*}{\makecell[c]{RAF-DB\\(400 labels)}} & a face image of \{\}                   & 78.58±\footnotesize{1.43} \\
                                           & this is a face image of \{\}              & 77.97±\footnotesize{0.89} \\
                                           & a/an \{\} expression is shown in the image & 78.47±\footnotesize{1.17} \\
        \midrule
        \multirow{3}{*}{\makecell[c]{SFEW\\(400 labels)}} & a face image of \{\}                      & 55.76±\footnotesize{0.81} \\
                                           & this is a face image of \{\}              & 55.16±\footnotesize{0.75} \\
                                           & a/an \{\} expression is shown in the image & 55.23±\footnotesize{0.48} \\
        \bottomrule
        \end{tabular}
    \label{evlText}
    \vspace{-0.2cm}
\end{table}

\vspace{0.3em}
\textbf{Evaluation of the Parameter \(\alpha\).} The parameter \(\alpha\) is the confidence threshold to filter out samples for learning where the maximum pseudo-label class probability exceeds the confidence threshold. We investigate its effect under values in $[0.6, 0.9]$. Figure \ref{fig:parameter}(a) shows that the performance is positively correlated with the increasing \(\alpha\) from 0.6 to 0.8. When \(\alpha\) exceeds 0.8, the performance degrades. This phenomenon occurs because an excessively large or small \(\alpha\) can respectively direct the model with either wrong ambiguous data or clear data. 

\vspace{0.3em}
\textbf{Evaluation of the Parameter \(\gamma\).} The parameter \(\gamma\) serves as a confidence threshold in deciding whether unlabeled data should be incorporated into the Instance Memory Buffer. We examined its impact by experimenting with values within the range of $[0.8, 0.9]$. As depicted in Figure \ref{fig:parameter}(b), the optimal result is achieved at $\gamma=0.86$. When the value of \(\gamma\) is set too low, an excessive amount of data with inaccurate pseudo-labels is deemed reliable, consequently diminishing the precision of instances in the Instance Memory Buffer. When the value of \(\gamma\) is set too high, some reliable data are considered to be unreliable. This leads to fewer updates in the buffer and fewer reference features.

\vspace{0.3em}
\textbf{Evaluation of different text contents.} In our SIT-FER framework, we introduce text contents to mitigate the challenge of unreliable pseudo-labels.  This is achieved by computing cosine similarity between image and text features, which is a pivotal component in our method. To rigorously evaluate the robustness of the SIT-FER model against variations in textual content, we undertake systematic comparative analyses, encompassing distinct text contents, including case 1: using a phrase in the format of “a face image of \{\}”, case 2: using a full active sentence “this is a face image of \{\}”, and case 3: using a passive sentence “a/an \{\} expression is shown in the image”. The detailed results are reported in Table \ref{evlText}. The phrase "a face image of \{\}" shows the best performance and there is little difference in model performance of different text content, indicating good robustness.

\subsection{Visualization}
\label{visualization}
\begin{figure}[t]
  \centering
   \includegraphics[width=1.00\linewidth]{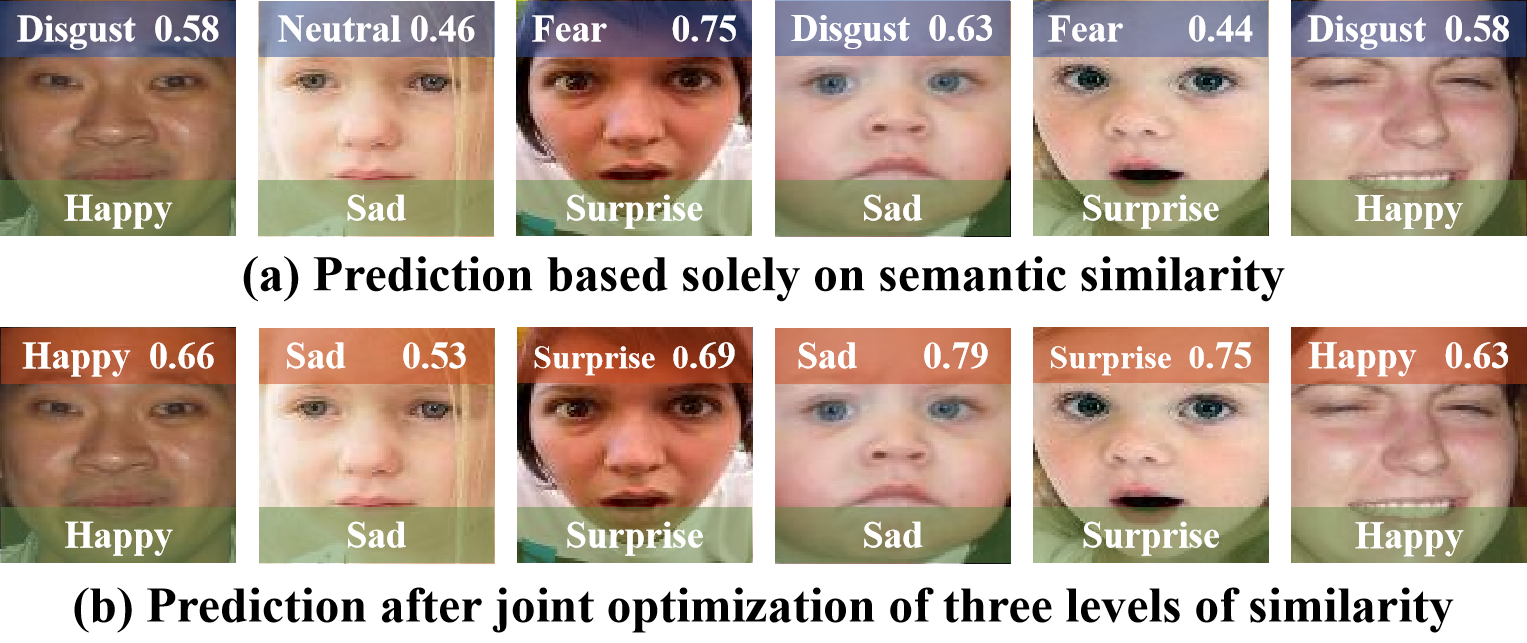}
   \caption{Visualization of the superiority of the three-level probability within our SIT-FER. The boxes at the top of the image display the predicted labels along with their respective probabilities, while the ground truth are presented at the bottom. Clearly, using three-level probability for prediction improves the accuracy.}
   \label{fig:visual}
\end{figure}

In this section, we further highlight the advantages of the pseudo-labels in SIT-FER that integrates three-level probabilities. Figure \ref{fig:visual} illustrates the comparison between traditional semantic-level pseudo-labels and our three-level pseudo-labels. As shown in Figure \ref{fig:visual}(a), we observe that relying solely on semantic-level information is not always reliable. After incorporating information from more levels, as illustrated in Figure \ref{fig:visual}(b), corrected pseudo-labels are closer to the ground truth, correcting erroneous semantic information. Take the first column as an example. In Figure \ref{fig:visual}(a), the predicted label is ‘Disgust’ with probability 0.58, while the ground truth is 'Happy'. In Figure \ref{fig:visual}(b), after applying the proposed three-level probability, the predicted label is corrected to 'Happy', which is the same as the ground truth, with probability 0.66.

\section{Conclusion}
\label{sec:conclusion}

In this paper, we propose a novel SS-DFER approach SIT-FER that simultaneously incorporates semantic-, instance-, and text-level information to generate high-quality pseudo-labels. In addition, to better leverage the one-hot labels, text embeddings excavated from textual descriptions are utilized for co-supervising model training. Extensive experiments on three datasets show that SIT-FER achieves state-of-the-art results and surpasses fully-supervised baselines.


\bibliographystyle{IEEEtran}
\bibliography{e}

\newpage

\vspace{11pt}

\begin{IEEEbiography}
[{\includegraphics[width=1in,height=1.25in,clip,keepaspectratio]
{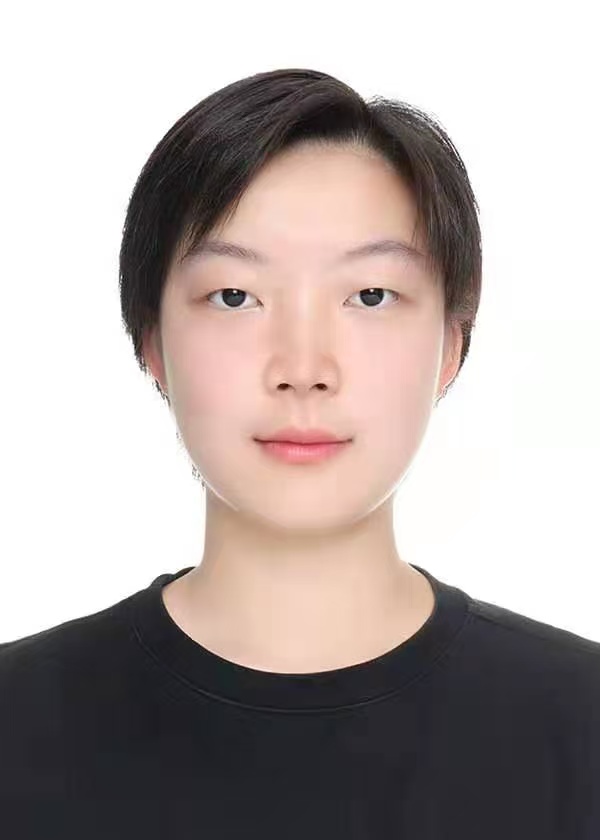}}]
{Sixian Ding} is currently pursuing a B.Eng. degree in Software Engineering at Sichuan University in Chengdu, China, under the guidance of Prof.Yan Wang. Her current research interests include image generation, image denoising, and facial expression recognition.
\end{IEEEbiography}

\begin{IEEEbiography}
[{\includegraphics[width=1in,height=1.25in,clip,keepaspectratio]
{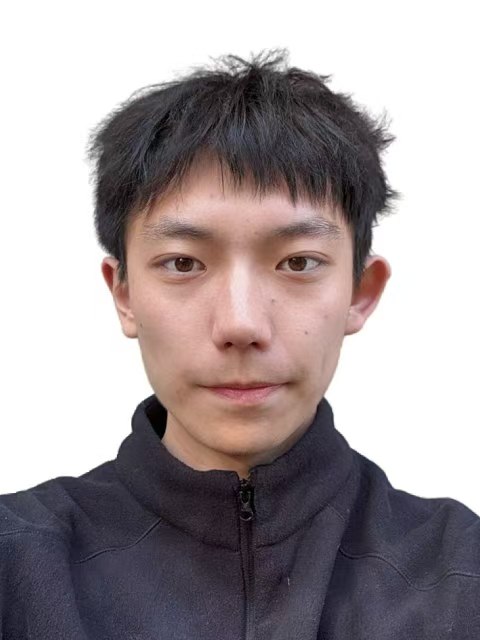}}]
{Xu Jiang} is currently pursuing a Bachelor's degree at the School of Software Engineering, Sichuan University, Chengdu, China. His research interests include multimodal large language model, image restoration, and facial expression recognition.
\end{IEEEbiography}

\begin{IEEEbiography}
[{\includegraphics[width=1in,height=1.25in,clip,keepaspectratio]
{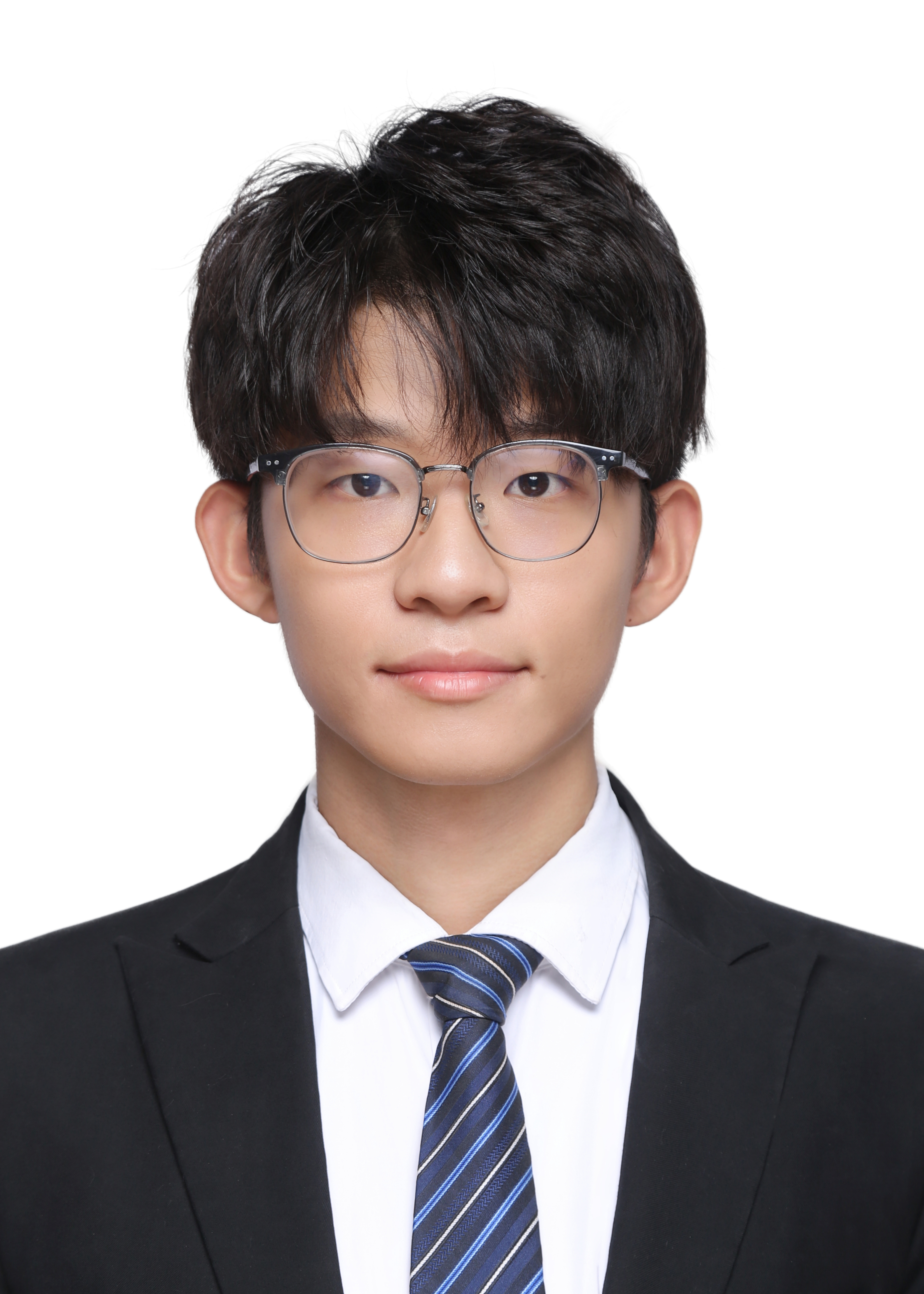}}]
{Zhongjing Du} is currently an undergraduate student at the School of Software Engineering, Sichuan University (Chengdu, China). His research interests include text-to-image generation and facial expression recognition.
\end{IEEEbiography}

\begin{IEEEbiographynophoto}
{Jiaqi Cui} is a doctoral student in School of Computer Science, Sichuan University. Her research interests include multimodal learning, medical image segmentation, medical image generation and PET image reconstruction.
\end{IEEEbiographynophoto}

\begin{IEEEbiographynophoto}
{Xinyi Zeng} is a doctoral student in School of Computer Science, Sichuan University. His research interests include deep learning, medical image analysis and domain adaptation.
\end{IEEEbiographynophoto}

\begin{IEEEbiography}
[{\includegraphics[width=1in,height=1.25in,clip,keepaspectratio]
{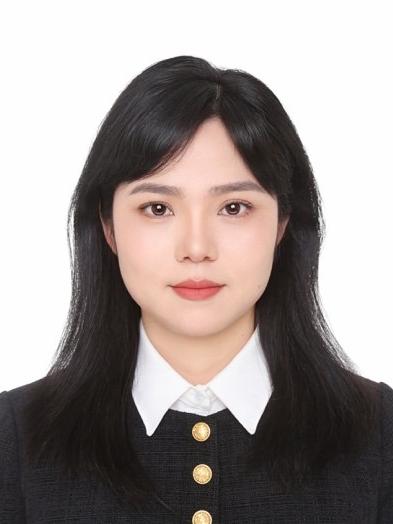}}]
{Yan Wang} received the Ph.D. degree from Sichuan University in 2015. She is currently a Professor at School of Computer Science, Sichuan University. She studied at University of North Carolina at Chapel Hill, USA and University of Wollongong, Australia as joint training Ph.D. student and Post doctorate in 2014–2015 and 2017–2018, respectively. Her research interests include computer vision, machine learning, deep learning, medical image analysis. 
\end{IEEEbiography}

\vfill

\end{document}